\documentclass[conference]{IEEEtran}
\usepackage{times}

\usepackage[sort,numbers]{natbib}
\usepackage{multicol}
\usepackage[bookmarks=true]{hyperref}
\usepackage{siunitx}
\usepackage{dirtytalk}
\usepackage{moreverb,url}
\usepackage{xspace}
\usepackage{xcolor}

\usepackage{amssymb}
\usepackage{amsmath}
\usepackage{amsthm}
\usepackage{amsbsy}
\usepackage{bbm}
\usepackage{dsfont}

\usepackage{graphicx}
\usepackage{wrapfig}
\usepackage{booktabs} 

\usepackage{subfiles}
\usepackage{csquotes}

\usepackage{array}

\usepackage{graphicx}
\usepackage{subfig} 
\usepackage{multirow}
\newcommand\BibTeX{{\rmfamily B\kern-.05em \textsc{i\kern-.025em b}\kern-.08em
T\kern-.1667em\lower.7ex\hbox{E}\kern-.125emX}}

\definecolor{wine}{RGB}{204, 0, 102}
\definecolor{magenta_wine}{RGB}{158, 44, 143}
\definecolor{dusty_wine}{RGB}{143, 59, 101}
\definecolor{ocean}{RGB}{13, 121, 202}
\definecolor{light_ocean}{RGB}{18, 178, 235}
\definecolor{dark_ocean}{RGB}{10, 89, 148}
\definecolor{grey}{RGB}{170, 170, 170}
\definecolor{light-grey}{RGB}{220, 220, 220}
\definecolor{dark_gray}{rgb}{0.2, 0.2, 0.2} 
\definecolor{med-grey}{rgb}{0.3, 0.3, 0.3} 
\definecolor{grape}{RGB}{112,48,160}
\definecolor{aqua}{RGB}{52,172,139}
\definecolor{dark_aqua}{RGB}{35,115,93}
\definecolor{dark_orange}{RGB}{216,92,0}
\definecolor{vibrant_orange}{RGB}{250, 160, 26}
\definecolor{vibrant_blue}{RGB}{14, 120, 255}
\definecolor{vibrant_pink}{RGB}{255, 0, 104}
\definecolor{dark_red}{RGB}{122, 0, 0}
\definecolor{dark_green}{RGB}{0, 92, 34}
\definecolor{dusty_blue}{RGB}{77, 91, 128}
\definecolor{dark_brown}{RGB}{125, 54, 36}

\theoremstyle{subtle} 

\newcommand{\para}[1]{\medskip\noindent\textbf{#1. }} 
 
\newcounter{qnum}
\newcommand{\ques}[1]{\medskip \noindent \textbf{Q\theqnum:} \textit{#1} \smallskip \stepcounter{qnum}}

\newcommand{\new}[1]{{#1}}

\newcommand{\ours}{\textcolor{vibrant_orange}{\textbf{LatentSafe}}\xspace}
\newcommand{\privsafe}{\textcolor{dark_brown}{\textbf{PrivilegedSafe}}\xspace}  
\newcommand{\cmdp}{\textcolor{ocean}{\textbf{SQRL}}\xspace} 
\newcommand{\dreamer}{\textcolor{dark_aqua}{\textbf{Dreamer}}\xspace} 
\newcommand{\diffpolicyadv}{\textcolor{magenta_wine}{\textbf{DiffusionAdv}}\xspace}
\newcommand{\diffpolicyopt}{\textcolor{grape}{\textbf{DiffusionOpt}}\xspace}

\newcommand{\env}{\mathbf{E}}
\newcommand{\envSpace}{\mathbb{E}}

\newcommand{\state}{s}
\newcommand{\stateSpace}{\mathcal{S}}
\newcommand{\dyns}{f}

\newcommand{\latent}{z}
\newcommand{\latentSpace}{\mathcal{Z}}

\newcommand{\encparam}{\psi}
\newcommand{\ellparam}{\mu}
\newcommand{\dynparam}{\phi}

\newcommand{\obs}{o}
\newcommand{\obsSpace}{\mathcal{O}}
\newcommand{\enc}{\mathcal{E}}
\newcommand{\img}{\mathcal{I}}

\newcommand{\action}{a}
\newcommand{\actionSpace}{\mathcal{A}}

\newcommand{\policy}{\pi}

\newcommand{\policyTask}{\pi^{\text{task}}}

\usepackage{fontawesome5}
\newcommand{\shield}{\text{\tiny{\faShield*}}}

\newcommand{\failure}{\mathcal{F}}
\newcommand{\failureLatent}{\failure_\textrm{latent}}
\newcommand{\marginfunc}{\ell}

\newcommand{\valfunc}{V}
\newcommand{\valfuncGt}{\valfunc_\textrm{gt}}
\newcommand{\valfuncLatent}{\valfunc_{\textrm{latent}}}
\newcommand{\valfuncPriv}{\valfunc_{\textrm{priv}}}

\newcommand{\unsafeSet}{\mathcal{U}} 

\newcommand{\fallback}{\policy^{\shield}} 

\newcommand{\fallbackGt}{\policy_\textrm{gt}^{\shield}} 
\newcommand{\fallbackLatent}{\policy^{\shield}_{\textrm{latent}}} 

\newcommand{\riskLatent}{\policy^{\textrm{risk}}_{\textrm{latent}}} 
\newcommand{\valfuncRisk}{\valfunc^{\mathrm{risk}}}

\newcommand{\fallbackPriv}{\policy^{\shield}_{\textrm{priv}}} 

\newcommand{\unsafeSetLatent}{\mathcal{U}_{\textrm{latent}}} 

\begin{document}

\title{Generalizing Safety Beyond Collision-Avoidance via Latent-Space Reachability Analysis}


\author{
Kensuke Nakamura \\
Carnegie Mellon University\\
kensuken@andrew.cmu.edu
\and
Lasse Peters \\
Delft University of Technology \\
l.peters@tudelft.nl
\and
Andrea Bajcsy \\ 
Carnegie Mellon University \\
abajcsy@andrew.cmu.edu
\vspace{-0.1em}
}

\makeatletter
\let\@oldmaketitle\@maketitle
\renewcommand{\@maketitle}{\@oldmaketitle
\setcounter{figure}{0} 
\centering
\includegraphics[width=0.99\textwidth]{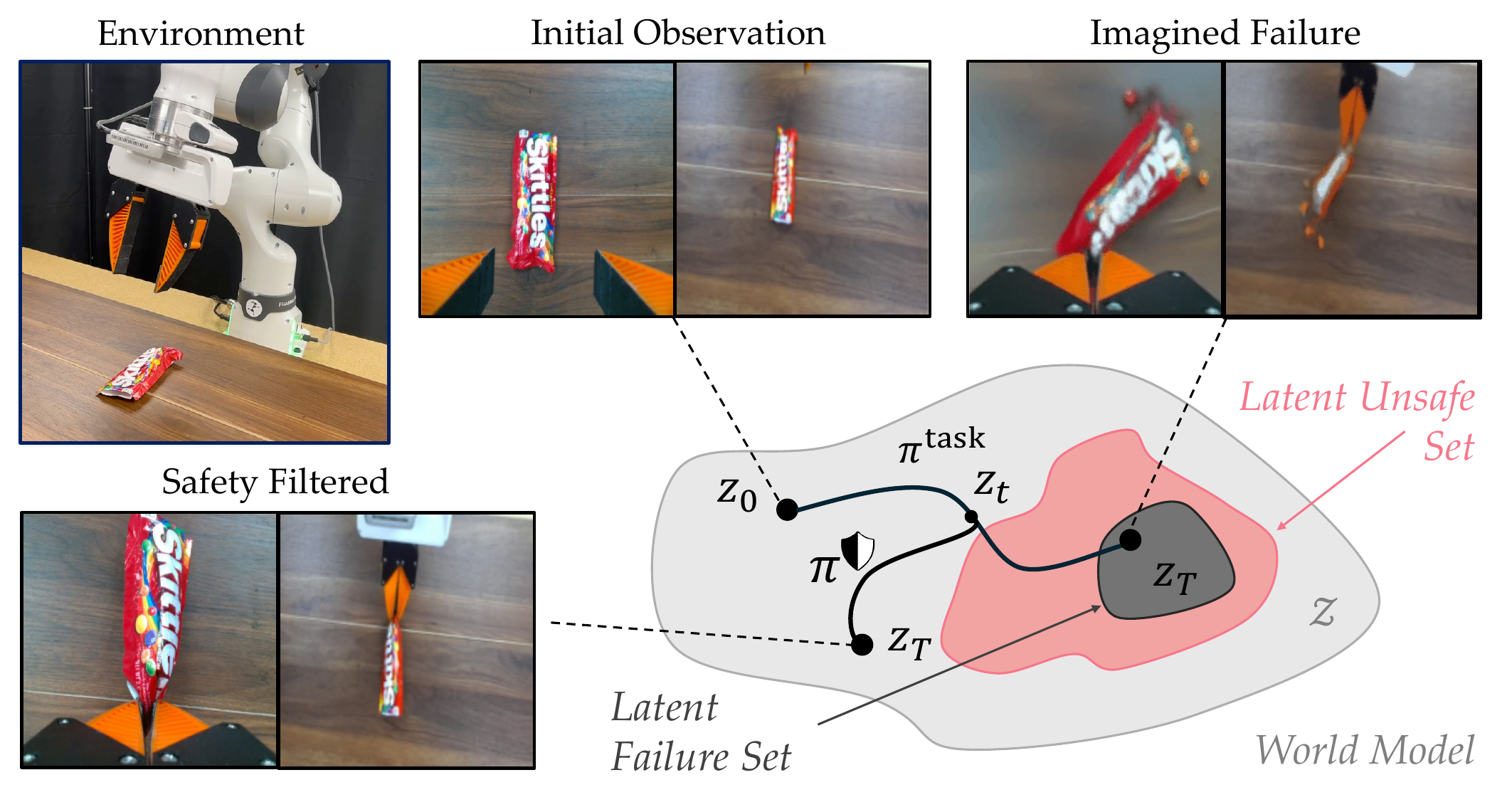}
\vspace{-1em}
\captionof{figure}{
Our \textit{Latent Safety Filter} can detect, predict, and mitigate failures that are hard to model (e.g., spilling the contents of a bag), such as those encountered in vision-based manipulation. 
Our idea is to perform approximate reachability analysis in the latent space of a world model (light grey region). 
The latent failure set is shown as a black region, with an example of an imagined failure observation shown in the upper right. 
Our method identifies latent states from which the robot is doomed to enter visually-observable failures no matter what actions it takes (larger red set shown above), and automatically overrides
the base policy $\policyTask$ with safety-preserving actions from our safety policy $\fallback$ to prevent spilling the content of the bag. \\ Video results can be found 
on the project website: \href{https://kensukenk.github.io/latent-safety/}{https://kensukenk.github.io/latent-safety/
}.
}
\label{fig:front-fig}
\vspace{-0.3in}
\bigskip}
\makeatother
\maketitle


\begin{abstract}

Hamilton-Jacobi (HJ) reachability is a rigorous mathematical framework that enables robots to simultaneously detect unsafe states and generate actions that prevent future failures. 
While in theory, HJ reachability can synthesize safe controllers for nonlinear systems and nonconvex constraints, in practice, it has been limited to hand-engineered collision-avoidance constraints modeled via low-dimensional state-space representations and first-principles dynamics. 
In this work, our goal is to \textit{generalize} safe robot controllers to prevent failures that are hard---if not impossible---to write down by hand, but can be intuitively identified from high-dimensional observations: for example, spilling the contents of a bag.  
We propose \textit{Latent Safety Filters}, a latent-space generalization of HJ reachability 
that tractably operates directly on raw observation data (e.g., RGB images) \new{to automatically computes safety-preserving actions without explicit recovery demonstrations} by performing safety analysis in the latent embedding space of a generative world model. \new{Our method leverages 
diverse robot observation-action data of varying quality (including successes, random exploration, and unsafe demonstrations) to learn a world model.  Constraint specification is then transformed into a classification problem in the latent space of the learned world model.}
In simulation and hardware experiments, we \new{compute an approximation of} Latent Safety Filters to safeguard arbitrary policies (from \new{imitation-learned} policies to direct teleoperation) from complex safety hazards, like preventing a Franka Research 3 manipulator from spilling the contents of a bag 
or toppling cluttered objects. 


%
%



\end{abstract}

\IEEEpeerreviewmaketitle

\section{Introduction}
Imagine that a robot manipulator is deployed in your home, like shown in Figure~\ref{fig:front-fig}. 
What safety constraints should the robot reason about? 
It is common to equate robot safety with \say{collision avoidance}, but in unstructured open world environments, a robot's representation of safety should be much more nuanced. For example, the household manipulator should understand that pouring coffee too fast will cause the liquid to overflow; pulling a mug too quickly from a cupboard will cause other dishes to fall and break; or, in Figure~\ref{fig:front-fig}, aggressively pulling up from the bottom of an open bag will cause the contents to spill. 


While these failure states---like liquid overflowing, objects breaking, items spilling---are possible to directly identify from high-dimensional observations, understanding how the robot could enter those states is extremely hard. 
Consider the set of failure states visualized in the black region in Figure~\ref{fig:front-fig}. These failures correspond to states where the Skittles have \textit{already} spilled onto the table. 
But even before the spill is visible, there are states from which the robot manipulator is doomed to end up spilling no matter what it does (visualized as a red set in Figure~\ref{fig:front-fig}): for example, after yanking the bottom of the bag up too quickly, no matter if the robot slows down or reorients the bag, the Skittles are doomed to spill all the way out.



This is where safe control theory can provide some insight. 
Frameworks like Hamilton-Jacobi (HJ) reachability analysis \cite{mitchell2005time, lygeros2004reachability} mathematically model safety constraints very generally---as arbitrary (non-convex) sets in a state space---and automatically identify states from which the robot is doomed to fail in the future by solving an optimal control problem. 
However, the question remains how to practically instantiate this theoretical framework to safeguard against more nuanced failures---\emph{beyond collision-avoidance}---in robotics.

Our key insight is that the latent representations learned by generative world models \cite{hafner2020dreamerv2, zhou2024dinowm} enable safe control for constraints not traditionally expressable in handcrafted state-space representations. 
\new{While world models require diverse coverage (success, play, and/or failure data) to accurately predict the dynamical consequences of robot actions,}
this formulation makes constraint specification as easy as learning a classifier in the latent space \new{\cite{wilcox2022ls3}}, and HJ reachability can safely evaluate possible outcomes of different actions (to determine if the robot \new{will inevitably} fail) within the ``imagination'' of the world model \new{without additional unsafe environment interactions}. 


We evaluate our approach in three vision-based safe-control tasks in both simulation and hardware: 
\begin{enumerate}
    \item a classic collision-avoidance navigation problem where we can compare our latent approximation with a privileged state solution, 
    \item a high-fidelity simulation of manipulation in clutter where the robot can touch, push, and tilt objects as long as they do not topple over, and 
    \item hardware experiments with a Franka Research 3
       arm picking up an open bag of Skittles without spilling. 
\end{enumerate}
Our quantitative results show that, without assuming access to ground-truth dynamics or hand-designed failure specifications, \textit{Latent Safety Filters} can learn a high-quality safety monitor ($F_1 \text{~score}: 0.984$) and the safety controller provides a $63.6\%$ safety failure violation decrease over a base policy trained via imitation \cite{chi2024diffusionpolicy}. 
In qualitative experiments, we also find that \textit{Latent Safety Filters} allow teleoperators to freely grasp, move, and pull up on an opened bag of Skittles, while automatically correcting any motions that would lead to spilling.

\new{\para{Statement of Contributions} To summarize, we make four key contributions in this work. 
\begin{itemize}
    \item  We formulate a HJ reachability problem in the latent space of a world model. This enables the specification of hard-to-model constraints that go beyond collision-avoidance (e.g., spilling) as a classification problem on the embedding space. The solution to our latent reachability problem automatically yields a safety-preserving policy and unsafe set that operate on high-dimensional RGB observations, without the need for explicit expert recovery demonstrations.
    \item We tractably compute \textit{Latent Safety Filters} via a reinforcement learning approximation \cite{fisac2019bridging} in the world model's latent ``imagination'', minimizing the need for additional unsafe online interaction data beyond what is needed to learn an accurate world model. 
    \item In a benchmark safe navigation task (with a privileged safety controller and unsafe set) and a contact-rich manipulation task in simulation, we find that \textit{Latent Safety Filters} steer a base policy away from failures while minimizing incompletion rates more effectively than soft constraints or constrained MDP formulations \cite{srinivasan2020learning, thananjeyan2021recovery}. 
    \item We deploy Latent Safety Filters in hardware where the constraint ``beyond collision-avoidance'' is not to spill Skittles from an opened bag during interaction with a Franka Panda 7DOF manipulator. Our experimental results demonstrate that the \textbf{same} \textit{Latent Safety Filter} minimally corrects a human teleoperator from spilling, does not impede a performant imitation-learned policy, makes a suboptimal imitation-learned policy safer, and generalizes to out-of-distribution Skittles bag colors and background changes.
\end{itemize}}

\section{A Brief Background on HJ Reachability}
\label{sec:hj-bg}

Hamilton-Jacobi (HJ) reachability \cite{mitchell2005time, margellos2011hamilton} is a control-theoretic safety framework for identifying when present actions will cause future failures, and for computing best-effort policies that minimize failures.  
Traditionally, reachability assumes access to a privileged state space $\state \in \stateSpace$ and a corresponding \new{bounded, discrete-time} nonlinear dynamics model $\state_{t+1} = \dyns(\state_t, \action_t)$\footnote{\new{Readers familiar with the \textit{continuous-time} HJ reachability formulation \cite{mitchell2005time} will note that a common assumption is Lipschitz-continuous dynamics: $\exists L \geq 0, ~ ||f(\state_1, \action) - f(\state_2, \action)|| \leq L ||\state_1 - \state_2||$. 
This ensures the existence and uniqueness of the state trajectory when the dynamics are an ordinary differential equation \cite{lygeros2004reachability}. 
In \textit{discrete-time}, assuming Lipschitz-continuous dynamics ensures convergence of traditional numerical methods for the non-discounted optimal control problem \cite{falcone1994discrete}. In this work, we only assume the dynamics are bounded, $\exists C \in \mathbb{R}, ~ ||f(\state, \action)|| \leq C$, as explained in Section~\ref{sec:lsf}}} which evolves via the robot's control action, $\action \in \actionSpace$ \new{where $\actionSpace$ is a compact set}.

A domain expert will first specify what safety means by imposing a constraint on the state space, referred to as the \textit{failure set}, $\failure \subset \stateSpace$. 
Given the failure set, HJ reachability will automatically compute two entities: (\romannumeral 1) a safety monitor, $\valfunc: \stateSpace \rightarrow \mathbb{R}$, which quantifies if the robot is doomed to enter $\failure$ from its current state $\state$ despite the robot's best efforts, and (\romannumeral 2) a best-effort safety-preserving policy, $\fallback: \stateSpace \rightarrow \actionSpace$. 
These two entities are co-optimized via the solution to an optimal control problem that satisfies the fixed-point safety Bellman equation \cite{fisac2019bridging}:
\begin{equation}
        \valfunc(\state) = \min \Big\{ \marginfunc(\state), ~ \max_{\action \in \actionSpace} \valfunc(\dyns(\state, \action)) \Big\},
        \label{eq:trad-fixed-pt-bellman}
\end{equation}
where $\marginfunc: \stateSpace \rightarrow \mathbb{R}$ \new{is a bounded margin function that} encodes the safety constraint $\failure$ via its zero-sublevel set $\failure = \{ \state \; | \; \marginfunc(\state) < 0 \}$, typically modeled as a signed-distance function.  
The maximally safety-preserving policy can be obtained via 
\begin{equation}
    \fallback(\state) := \arg\max_{\action \in \actionSpace} \valfunc(\dyns(\state, \action)).
\end{equation}
 Finally, the \textit{unsafe set}, $\unsafeSet \subset \stateSpace$, which models the set of states from which the robot is doomed to enter $\failure$, can be recovered from the zero-sublevel set of the value function: $\unsafeSet := \{\state : \valfunc(\state) < 0\}$.

At deployment time, the safety monitor and safety policy can be utilized together to perform \textit{safety filtering}: detecting an unsafe action generated by any base policy, $\policyTask$, and minimally modifying it to ensure safety.
While there are a myriad of safety filtering schemes (see surveys \cite{wabersich2023data, hsu2023safety} for details), a common minimally-invasive approach switches between the nominal and the safety policy when the robot is on the verge of being doomed to fail: $\action^\text{exec} = \mathds{1}_{\{\valfunc(\state) > 0\}} \cdot \policyTask + \mathds{1}_{\{\valfunc(\state) \le 0\}} \cdot \fallback$.



\section{Latent Safety Filters}
\label{sec:lsf}


To tackle both detecting and mitigating hard-to-model failures, we present a latent-space generalization of HJ reachability (from Section~\ref{sec:hj-bg}) that tractably operates on raw observation data (e.g., RGB images) by performing safety analysis in the latent embedding space of a generative world model.
This also transforms nuanced constraint specification into a classification problem in latent space and enables reasoning about dynamical consequences that are hard to simulate. 



\para{Setup: Environment and Latent World Models} We model the robot as operating in an environment $\env \in \envSpace$, which broadly characterizes the deployment context---e.g., in a manipulation setting, this includes the geometry and material properties of the table, objects, and gripper. The robot has a sensor $\sigma : \stateSpace \times \env \rightarrow \obsSpace$ that generates observations $\obs \in \obsSpace$
depending on the true state of the world.
While we are in a partially-observable setting and never have access to the true state, we will leverage a world model to jointly infer a lower-dimensional latent state and its associated dynamics that correspond to the high-dimensional observations.

A world model consists of an encoder that maps observations $\obs_t$ (e.g., images, proprioception, etc.) and  latent state $\hat{\latent}_t$ into a posterior latent $\latent_t$,
and a transition function that predicts the future latent state conditioned on an action. This can be mathematically described as:

\begin{itemize}
    \smallskip  
    \item[] Encoder: $\latent_t \sim \enc_\encparam(\latent_t \mid \hat{\latent}_t, \obs_t)$
    \item[] Transition Model: $\hat{\latent}_{t+1} \sim p_\dynparam(\hat{\latent}_{t+1} \mid \latent_t, \action_t)$. 
    \smallskip  
\end{itemize}
This formulation describes a wide range of world models~\cite{ha2018recurrent, hafner2019learning, hafner2020dreamerv2, hafner2024masteringdiversedomainsworld,zhou2024dinowm}, and our latent safety filter is not tied to a particular world model architecture. 
We focus on world models that are trained via self-supervised learning (observation reconstruction, teacher forcing, etc.) and do not require access to a privileged state. 
Specifically, in Section \ref{sec:sim} we use a Recurrent State Space Model (RSSM) \cite{hafner2019learning} trained with an observation reconstruction objective and in Section \ref{sec:hardware} we use DINO-WM \cite{zhou2024dinowm} which is trained via teacher-forcing.

\para{Safety Specification: Failure Classifier on Latent State} A common approach for representing $\failure$ is to encode it as the zero-sublevel set of a function $\marginfunc(\state)$ (as in Eq.~\ref{eq:trad-fixed-pt-bellman}). Domain experts typically design this ``margin function'' to be a signed distance function to the failure set, which easily expresses constraints like collision-avoidance (e.g., distance between positional states of the robot and environment entities being less than some threshold). 
However, other types of constraints, such as liquid spills, are much more difficult to directly express with this class of functions and traditional state spaces. 
To address this, we choose to learn $\marginfunc_\ellparam(\latent)$ from data by modeling it as a classifier over latent states~$\latent\in\latentSpace$, with learnable parameters $\ellparam$.

We train our classifier on labeled datasets of observations corresponding to safe and unsafe states, $\obs^+ \in \mathcal{D}_{\text{safe}}$ and $\obs^- \in \mathcal{D}_{\text{unsafe}}$, and optimize the following loss function inspired by~\cite{yusequential}:
\begin{equation}
\begin{aligned}
     \mathcal{L}(\ellparam) = &\frac{1}{N_\text{safe}}\sum_{\obs^+ \in \mathcal{D}_{\text{safe}}} \text{ReLU}\big(\delta - \marginfunc_\ellparam(\enc_\encparam(\obs^+))\big)\\ + &\frac{1}{N_\text{fail}}\sum_{\obs^- \in \mathcal{D}_\text{fail}} \text{ReLU}\big(\delta + \marginfunc_\ellparam(\enc_\encparam(\obs^-))\big), 
     \label{eq:failure-classifier-loss}
\end{aligned}
\end{equation}
where the loss function is parameterized by $\delta \in \mathbb{R}^+$ to prevent degenerate solutions where all latent states are labeled as zero by the classifier.
Intuitively, this loss penalizes latent states corresponding to observations in the failure set from being labeled positive and vice versa. 
The learned classifier represents the failure set $\failureLatent$ in the latent space of the world model via: $\failureLatent = \{\latent \mid \marginfunc_\ellparam(\latent) < 0 \}$.  Our failure classifier can be co-trained (Section \ref{sec:sim}) or trained after (Section \ref{sec:hardware}) world model learning.

\para{Latent-Space Reachability in Imagination} Traditionally, reachability analysis requires either an analytic model of the robot and environment dynamics \cite{mitchell2004toolbox, bansal2021deepreach} or a high-fidelity simulator \cite{fisac2019bridging, hsu2023isaacs} to solve the fixed-point Bellman equation, both of which are currently inadequate for complex system dynamics underlying nuanced safety problems (e.g., liquid interaction). 
Instead, we propose using the latent imagination of a pretrained world model as our environment model, capturing  hard-to-design and hard-to-simulate interaction dynamics. 
We introduce the latent fixed-point Bellman equation: 
\begin{equation}
    \valfuncLatent(\latent) = \min \Big\{ \marginfunc_\ellparam(\latent), ~ \max_{\action \in \actionSpace} \mathbb{E}_{\hat{\latent}' \sim p_\dynparam(\cdot \; | \;\latent, a)} \Big[\valfuncLatent(\hat{\latent}') \Big]\Big\}.
    \label{eq:latent-fixed-pt-bellman}
\end{equation}
Note that in contrast to Equation~\ref{eq:trad-fixed-pt-bellman}, this backup operates on the latent state $\latent$ and, for full generality, includes an expectation\footnote{This expectation could also be taken to be a risk metric (e.g., CVaR) or a worst-over-N samples of the transition function to induce additional conservativeness.} over transitions to account for world models with stochastic transitions (e.g., RSSMs). For world models with deterministic transitions (e.g., DINO-WM), the expectation can be removed. 


While the world model allows us to compress high-dimensional observations into a compact informative latent state, computing an exact solution to the latent reachability problem is still intractable due to the dimensionality of the latent embedding (e.g., our latent is a 544 dimensional vector in Section \ref{sec:sim}). This motivates the use of a learning-based approximation to the value function in Equation~\ref{eq:latent-fixed-pt-bellman}. We follow \cite{fisac2019bridging} and induce a contraction mapping for the Bellman backup by adding a time discounting factor $\gamma \in [0,1)$:
\begin{multline}
        \label{eqn:discounted-safety-bellman}
        \valfuncLatent(\latent) = (1-\gamma)\marginfunc_\ellparam(\latent) \\ + \gamma \min \Big\{ \marginfunc_\ellparam(\latent), \max_{\action \in \actionSpace}     \mathbb{E}_{\hat{\latent}' \sim p_\dynparam(\cdot \; | \;\latent, a)}[\valfuncLatent(\hat{\latent}')] \Big\}
\end{multline}
\new{We note that the contraction induced by this time-discounted Bellman backup converges to a unique value function under the mild assumptions on the boundedness of the margin function and dynamics \cite{Bertsekas_2018, bellman1952theory}.}

In theory, if solved to optimality, this latent value function would offer a safety assurance only with respect to 
the data used to train the world model and the failure classifier. 
Intuitively, this implies that the robot can only provide an assurance that it will try its hardest to avoid failure \textit{in its representation of the world}. 
In the following section, we study our overall latent safety framework and the effect of world model dataset coverage on a benchmark safe control task for which we have exact solutions. 
We then scale to high-dimensional manipulation examples in both simulation and hardware.

\section{Simulation Results}
\label{sec:sim}

We conduct simulation experiments across two different vision-based tasks to assess the performance of our latent safety framework. 
These experiments are designed to support the claim that latent safety filters recover performant safety-preserving policies from partial observations alone (i.e., without assuming access to ground-truth dynamics, states, or constraints), for progressively more complex safety specifications and dynamical systems.

\subsection{How Close Does Latent Safety Get to Privileged Safety?}
We start by studying a canonical safe-control benchmark: collision-avoidance of a static obstacle with a vehicle.
Although this particular setting does not ``require'' latent-space generalizations of safety, its low-dimensionality and well-studied unsafe set allow us to 
rigorously compare the quality of the safety filter to a \new{traditional numerical} grid-based solution \cite{bui2022optimizeddp} \new{which we take as ground-truth} and 
a privileged-state RL-based safety filter.



\begin{figure}[t!]
    \centering
    \includegraphics[width=1\linewidth]{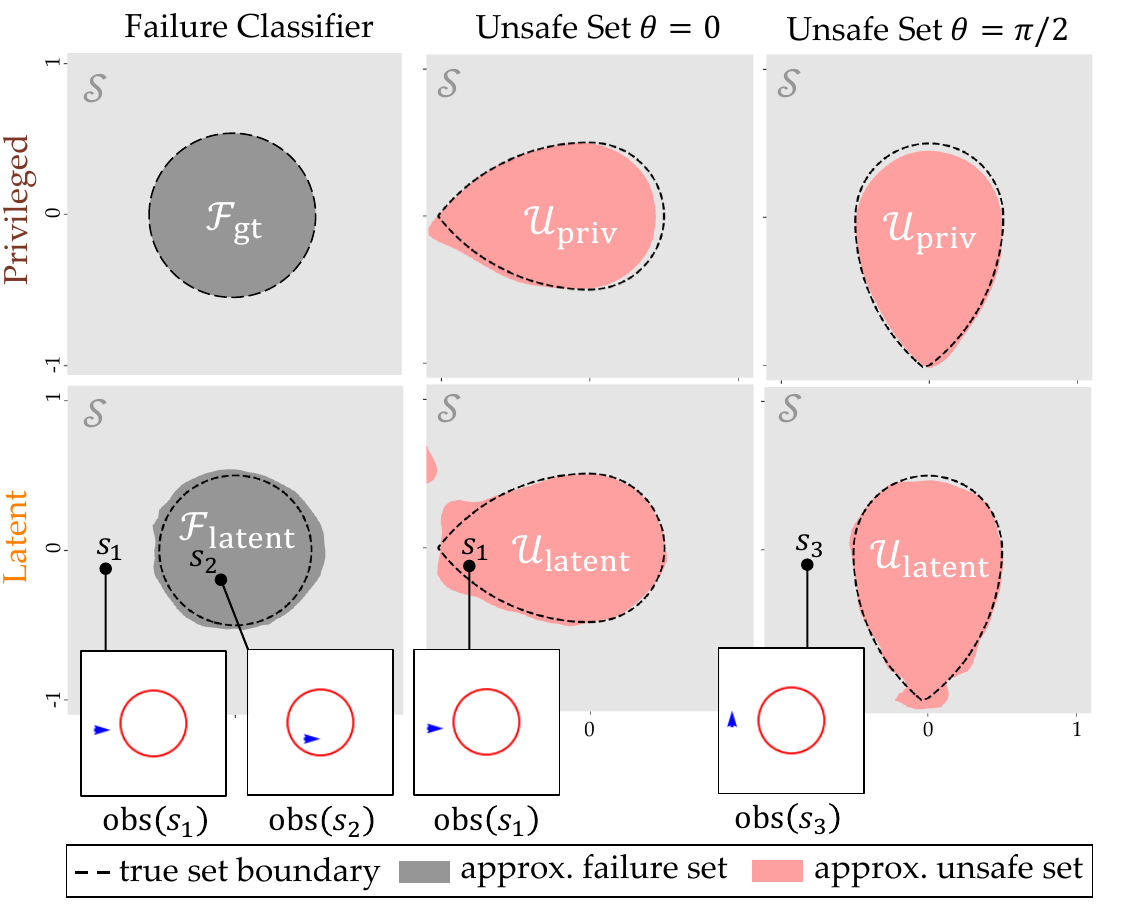}
    \caption{\textbf{Latent Safety vs. Privileged Safety.} Dubins' car collision-avoidance qualitative results. Dashed lines indicate the ground-truth set boundary. We visualize each method's failure specification and corresponding unsafe set, shown at heading slices $\theta \in \{ 0, \pi/2 \}$. While \privsafe uses the ground-truth $\state$ and $\failure_\textrm{gt}$, \ours uses the latent state from encoding the observation, $\latent = \enc_\encparam(\obs)$, and the inferred failure set $\failureLatent$. Insets on the bottom row show the observations corresponding to select privileged states $\state_1, \state_2, \state_3.$ }
    \label{fig:dubins-results}
\end{figure}

\para{Dynamical System \& Safety Specification} 
In this experiment, the ground-truth dynamics are that of a discrete-time 3D Dubin's car with state $\state = [p^x,p^y,\theta]$ evolving as
$$\state_{t+1} = f(\state_t, \action_t) 
= \state_t + \Delta t [v \cos(\theta_t), v \sin(\theta_t), \action_t],$$ 
where the robot's action  $\action$ controls angular velocity while the longitudinal velocity is fixed at $v = \SI{1}{\meter\per\second}$, and the time-discretization is $\Delta t = \SI{0.05}{\second}$.
The action space for the robot is discrete and consists of $\mathcal{A} = \{-a_\mathrm{max}, 0, a_\mathrm{max} \}$ where $a_\mathrm{max} = 1.25~\mathrm{rad/s}$.  
To remain safe, the robot must avoid an obstacle of radius $r = \SI{0.5}{\meter}$ centered at the origin (see red circle in the bottom row of Figure~\ref{fig:dubins-results}).
Hence, the ground-truth failure set is a cylinder in state space, $\failure_\textrm{gt} := \{ \state : \forall \theta, \; |p^x|^2 + |p^y|^2 < r^2\}$.
The ground-truth margin function $\marginfunc_\textrm{gt}(\state) = |p^x|^2 + |p^y|^2 - 0.5^2$ captures the failure condition via its zero-sublevel set.  

\para{Baseline: Privileged Safety} The \privsafe baseline computes the HJ value function using the \emph{ground-truth} state, dynamics, and margin function, using a reinforcement learning (RL)-based solver.
Specifically, we approximate solutions to the discounted Bellman equation~(\ref{eqn:discounted-safety-bellman}) in the framework of~\cite{hsu2021safety} via DDQN~\cite{van2016deep} with Q-functions parameterized by a 3-layer MLP with 100 hidden units. \new{We evaluate $\pi^\shield_\textrm{priv}(\state) = \arg\max_{\action \in \actionSpace} Q^\shield_\textrm{priv}(\state, \action)$ via action enumeration and a simple lookup procedure due to the discrete action space.}

\para{Latent Safety Filter Setup}
The \ours method does not get privileged access to ground-truth information but instead learns all model components from data.
Specifically, we train a world model from an offline dataset $\mathcal{D}_\mathrm{WM} = \{\{(o_t, a_t)\}_{t=0}^T\}^N_{i=1}$ of $N=2,000$ observation-action trajectories. The observations $\obs_t := (\img_t, \theta_t)$ consist of (128x128) RGB images $\img$ and the robot heading $\theta_t \in [-\pi, \pi]$. 
During trajectory generation, we uniformly sampled random actions. 
Each trajectory terminates after $T=100$ timesteps or if the ground-truth $x$ or $y$ coordinate left the environment bounds of $[\SI{-1}{\meter}, \SI{1}{\meter}]$. 
We trained the world model on this offline dataset using the default hyperparameters of~\cite{dreamerv3-torch}.
We used the privileged state to automatically label each observation $o_t \in \mathcal{D}_\mathrm{WM}$ as violating a constraint or not, constructing the datasets $\mathcal{D}_{\text{safe}}$ and $\mathcal{D}_{\text{unsafe}}$ for classifier training in Eq.~\ref{eq:failure-classifier-loss}. 
For \ours, we parameterize the safety classifier $\marginfunc_\ellparam(\latent)$ by a 2-layer MLP with 16 hidden units and co-train it with the world model.
The zero-sublevel set of $\marginfunc_\ellparam(\latent)$ captures the learned failure set, $\failureLatent := \{ \latent \mid \marginfunc_\ellparam(\latent) < 0\}$.

\new{One important implementation detail when solving Equation~\eqref{eqn:discounted-safety-bellman} via reinforcement learning in latent space is dealing with the reset mechanism in the world model's imagination.} 
Naively initializing the latent state may result in a latent state that does not correspond to any observation seen by the world model. 
In practice, we reset the world model by encoding a random observation from our offline dataset and only collect rollouts in the latent imagination for $T=25$ steps to prevent drifting too far out-of-distribution.
\new{For approximating the latent HJ value function and safety controller, we use the same toolbox and hyperparameters as the privileged baseline. 
We also obtain the safe control via the same procedure as for the priviledged state baseline, via enumerating over the discrete action space to solve for $\pi^\shield_\textrm{latent}(\latent) = \arg\max_{\action \in \actionSpace} Q^\shield_\textrm{latent}(\latent, \action)$}





\para{Results: On the Quality of the Runtime Monitor, $\valfuncLatent$}
Since the ground-truth dynamics are known and its state-space is low-dimensional, we can solve for the safety value function \emph{exactly} using traditional grid-based methods \cite{mitchell2004toolbox}.
This allows us to report the accuracy of the safe/unsafe classification of the safety filter's monitor ($\valfunc$) in Table \ref{tab:filter_table} for both the privileged state value function $\valfuncPriv$ and latent state value function $\valfuncLatent$ based on their alignment (in terms of sign) with the ground-truth value function $\valfuncGt$.
Since both safety filters use the same toolbox for learning the value function, any degradation in the latent safety filter can be attributed to the quality of the learned world model.
We also qualitatively visualize the zero-sublevel set of the value function for both methods at different fixed values of $\theta$ in Figure~\ref{fig:dubins-results}. 
In summary, we find that the \emph{image-based} runtime monitor learned by our method (\ours) closely matches the accuracy of the \emph{true-state-based} of the privileged baseline (\privsafe).

\begin{table}[h!]
\centering
\setlength{\tabcolsep}{2.5pt} 
\begin{tabular}{l|cccc|c}
\toprule
Method & True Safe & False Safe & False Unsafe & True Unsafe & $F_1$-score \\ 
\midrule
\privsafe & 0.771 & 0.011 & 0.008 & 0.209 & 0.987 \\ 
\ours     & 0.759 & 0.003 & 0.021 & 0.217 & 0.984 \\ 
\bottomrule
\end{tabular}
\caption{\textbf{Quality of the Runtime Monitor.} Performance of latent (\ours) and privileged (\privsafe) safety value functions. Note that this is computed over all three dimensions of the Dubins' car state.}
\label{tab:filter_table}
\end{table}


\para{Results: On the Quality of the Safety Policy, $\fallbackLatent$}
Each runtime monitor $\valfunc$ induces a corresponding safety policy: $\fallbackLatent$ for \ours and $\fallbackPriv$ for \privsafe.
To evaluate the quality of these safety policies, we check if they are capable of steering the robot away from failure. To determine initial conditions from which steering away from failure is feasible, we compute a state-based ground-truth value function, $\valfuncGt$, whose zero-sublevel set gives us a dense grid of 250 initial states for which the exact safe controller $\fallbackGt$ can guarantee safety.
For all of these states, we simulate each policy executing its best effort to keep the robot outside the privileged failure set, $\failure_\state$.
We find that our \emph{image-based} safety policy closely matches the performance of the \emph{privileged} baseline: \ours maintains safety for 240/250 (96\%) states and \privsafe maintains safety for 246/250 (98.4\%) states.

\begin{figure}[h!t]
    \centering
    \includegraphics[width=1\linewidth]{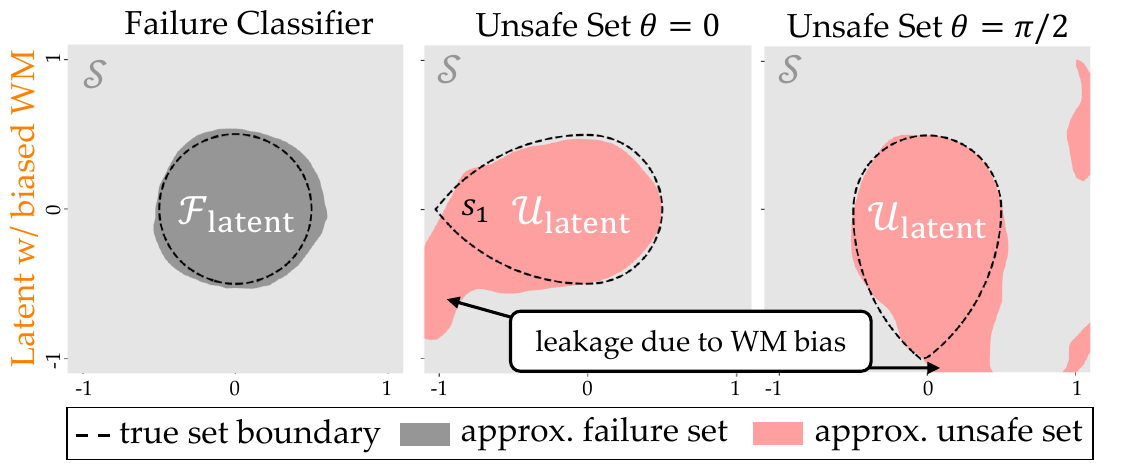}
    \caption{\textbf{Ablation: Latent Safety with Incomplete WM.}
    Unsafe set approximated by \ours using the latent space of a \emph{biased} world-model built from incomplete action coverage $\tilde{\actionSpace}=\{0, \action_\textrm{max}\}\subset \actionSpace$.
    }
    \label{fig:dubins-ablation}
\end{figure}

\para{Ablation: Effect of Incomplete Knowledge of the World}
Thus far, we have had strong coverage of all observation-action pairs when training the world model; however, complete knowledge of the world may not be achievable in reality.
To study this, we train our latent safety filter on top of a world model that has seen a biased dataset, wherein the robot's action space is limited to only moving straight or turning left: $\tilde{\actionSpace} = \{0, \action_\mathrm{max} \} \subset \actionSpace$.
Figure~\ref{fig:dubins-ablation} shows that the bias of the world model affects the robot's understanding of safety: since the world model did not learn about the possibility of turning right, \ours pessimistically classifies states as unsafe if they require a right turn to avoid collision. 

\subsection{Can Latent Safety Scale to Visual Manipulation?}

\begin{figure*}[t!]
    \centering
    \includegraphics[width=1\linewidth]{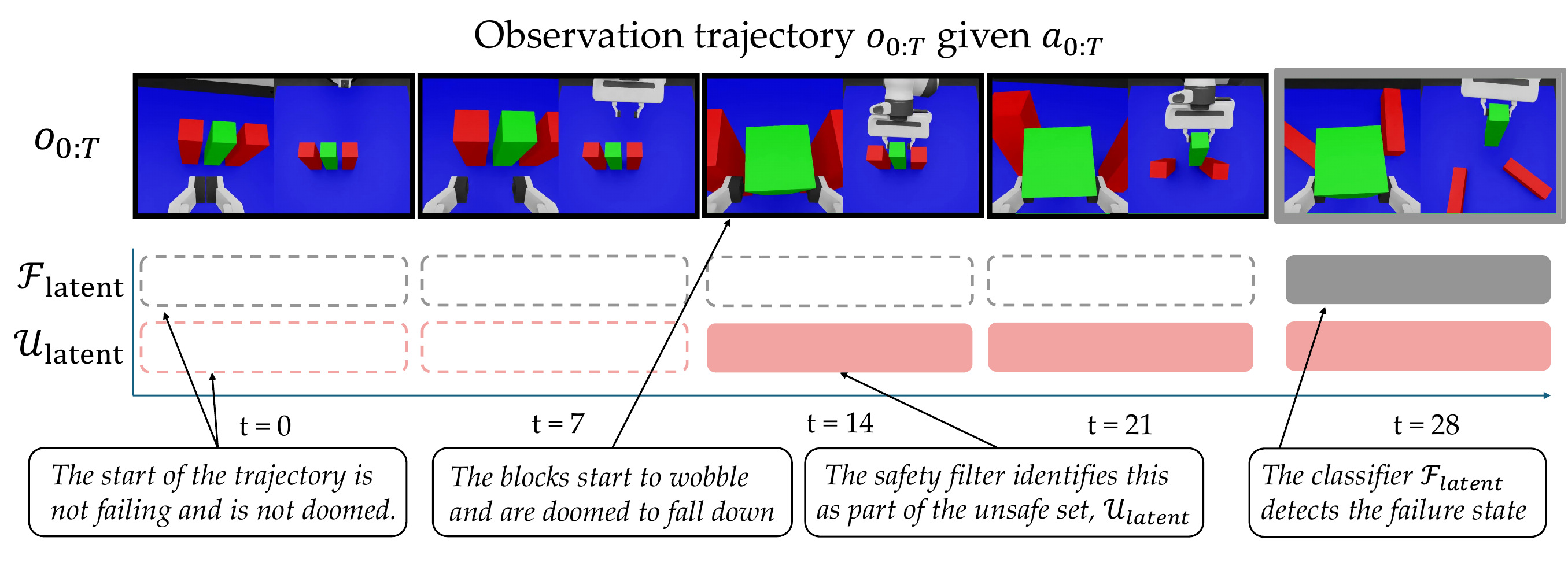} 
    \caption{
    \textbf{Visual Manipulation: Simulation.} \textit{Top row}: Robot's observations corresponding to a known \textit{unsafe} action sequence. \textit{Middle row}: Our learned failure classifier correctly identifies only the final observations at $t=28$ as being in the failure state since the red blocks have fallen all the way over. \textit{Bottom row}: Our unsafe set (obtained via the latent-space HJ value function) correctly identifies that the robot is doomed to fail the moment that the two red blocks begin to tip over at time $t=14$. 
    }
    \label{fig:sim-manip}
\end{figure*}


In our simulated manipulation setting, we adapt a contact-rich manipulation task from \cite{thananjeyan2021recovery} where a robot is tasked with grasping and lifting a green block that is placed closely between two red blocks (see Figure \ref{fig:sim-manip}).
In this setting, we generalize the safety representation to more nuanced failures, such as the red blocks falling down from aggressive interaction. 
We train a task policy for this setting that accounts for safety only via a soft constraint and compare the unfiltered behavior against two methods for safety filtering: a constrained MDP (CMDP) baseline and our latent safety filter.


\para{Safety Specification} 
We treat a state as a failure if either of the two red blocks is knocked down.
We categorize a block as having fallen if it is angled within \SI{1}{\radian} (measured using privileged simulator information not seen by any of the methods) of the ground plane. 
Crucially, this safety specification is \textit{not} a collision avoidance specification: the robot is allowed to touch, push, and tilt the red blocks in order to grasp the green block, as long as the red blocks do not topple over.

\para{Experimental Setup} 
Our nominal task policy $\policyTask$ is obtained via DreamerV3 \cite{hafner2024masteringdiversedomainsworld} trained using a dense reward for lifting the block and a sparse cost for violating constraints ($\dreamer$). 
The observation space $\obsSpace$ of the robot is given by two $3 \times 128 \times 128$ RGB camera views (table view and wrist-mounted) along with 8-dimensional proprioception (7-dimensional joint angle and gripper state) information. 
We co-train the $\dreamer$ world model with our failure classifier $\marginfunc_\ellparam(\latent)$ for 100k iterations and reuse the world model with frozen weights for our method and all other baselines. The failure classifier is implemented as a 3-layer MLP with ReLU activations. During training, we sampled batches consisting of both rollouts collected by the Dreamer policy (90$\%$ of each batch) and a dataset of 200 teleoperated demonstrations (10$\%$ of each batch) comprising both safe and unsafe behavior. 

We also compare \ours to a constrained MDP safety policy, \cmdp \cite{srinivasan2020learning, thananjeyan2021recovery}. 
This method leverages the same world model and failure classifier as \ours, but optimizes a different loss function to obtain the safety critic, $Q^\textrm{risk}$ \cite{srinivasan2020learning}:
\begin{equation}
\begin{aligned}
    \mathcal{L}(\theta) = &\mathbb{E}_{(\hat{\latent}_t, \action_t, \hat{\latent}_{t+1}, \action_{t+1
    }) \sim \rho_{\riskLatent}} \bigg[ \frac{1}{2}(Q^\textrm{risk}(\hat{\latent}_t, \action_t) \\&- (c_t + (1-c_t)\gamma Q^\textrm{risk}(\hat{\latent}_{t+1},\action_{t+1})))^2\bigg]
\end{aligned}
\end{equation}
where $\rho_{\riskLatent}$ is the state-action distribution induced by policy $\riskLatent$ and $c_t \in \{0, 1\}$ takes value $1$ if a constraint is violated at timestep $t$. 
The resulting Q-value can be interpreted as the empirical risk of violating a constraint by taking action $\action_t$ from $\latent_t$ and following policy $\riskLatent$ thereafter. 
To mimic the role of our safety filter that is agnostic to any task-driven base policy, we train the \cmdp critic with trajectories sampled from $\riskLatent(\latent) := \arg\min_{\action \in \actionSpace} Q^\textrm{risk}(\latent, \action)$ and no additional task-relevant information. 
We can then use this policy to filter any actions whose risk is higher than some threshold, $\epsilon^\textrm{risk}$. The actor and critic for both \cmdp and \ours are trained in the latent dynamics of the RSSM using DDPG~\cite{lillicrap2019continuouscontroldeepreinforcement, li2025certifiabledeeplearningreachability}. \new{Since we are in the continuous action space, DDPG trains both a critic network $\valfuncLatent(\latent)$ that approximates the safety value function, and an actor network $\pi_\textrm{latent}(\latent)$ which is trained to optimize $\valfuncLatent(p_\dynparam(\hat{\latent}'\mid \latent, \action))$ since a table-lookup is not possible in continuous action spaces}. 
We again reset the latent state of the world model by encoding an observation of previously collected data. All training hyperparameters are included in the Appendix. 

During deployment, we use the actor head of $\dreamer$ to be the nominal policy, $\policyTask(\latent)$. 
To get a performant nominal policy, we tuned the \dreamer reward function by sweeping over the reward weights for the components consisting of reaching the green block, lifting the green block, action regularization, and a penalty for the red blocks falling.

\para{Safety Filtering} 
We instantiate our two safety filters, \ours (comprised of $\fallbackLatent(\latent)$ and $\valfuncLatent(\latent)$) and \cmdp (comprised of $\riskLatent(\latent)$ and $\valfuncRisk(\latent)$), to shield the base \dreamer policy. 
During each timestep $t$, we query a candidate action $\action_t$ from \dreamer that we seek to filter. 
We instantiate a modified version of the minimally-invasive safety filtering scheme described at the end of Section~\ref{sec:hj-bg}. We take the action in the world model to obtain latent state $\hat{\latent}_{t+1}$ and evaluate this latent state to obtain $\valfunc(\hat{\latent}_{t+1})$.
This value $\valfunc(\hat{\latent}_{t+1})$ will serve as our monitoring signal for whether we are safe or if we should start applying our safety policy.
For \ours, the filtered (and thus executed) action $\action^\text{exec}_{t}$ follows the filtering law: 
\begin{align}
   \action^\text{exec}_{t} = 
    \begin{cases}
    \policyTask(\latent_t) ,& \text{if } \valfunc(\hat{\latent}_{t+1}) > \epsilon, \\
    \fallbackLatent(\latent_t),              & \text{otherwise.}  \\ 
    \end{cases}
\end{align}
This is a least-restrictive filter\footnote{\new{While in theory $\epsilon = 0$ is an appropriate threshold for safety filtering, practitioners often select $\epsilon > 0$ to account for potential numerical errors or latency in the system.}} that executes the safe control policy $\fallbackLatent$ whenever $\valfunc(\hat{\latent}_{t+1}) \leq \epsilon = 0.4$. 
The \cmdp baseline follows a similar filtering control law defined by:
\begin{align}
   \action^\text{exec}_{t} = 
    \begin{cases}
    \policyTask(\latent_t) ,& \text{if } V^{\textrm{risk}}(\hat{\latent}_{t+1}) < \epsilon_{\textrm{risk}}, \;\; \\
    \riskLatent(\latent_t),              & \text{otherwise.}  \\ 
    \end{cases}
\end{align}
where $\epsilon_{\textrm{risk}}$ is a manually tuned risk threshold that we ablate in our experiments to be $\epsilon_{\textrm{risk}} \in \{0.1, 0.05\}$.


\para{Results: Qualitative} 
First, we qualitatively studied if there was a difference between the failure set, $\failureLatent$, learned by our classifier and the unsafe set, $\unsafeSetLatent$, recovered by learning the HJ value function in this visual manipulation task. 
If $\unsafeSetLatent \supset \failureLatent$, then we have identified a non-trivial unsafe set for this high-dimensional problem. 
In Figure \ref{fig:sim-manip}, we show the observations $\obs_{0:T}$ corresponding to a known \textit{unsafe} action sequence, $\action_{0:T}$, where $T = 28$ steps. 
When the robot executes this action sequence, half-way through the robot touches the red blocks with high enough force that they end up falling over. 
We pass the observation trajectory into world model encoder to obtain a corresponding posterior latent state trajectory $\latent_{0:T}$. 
We evaluate $\text{sign}[\marginfunc_\ellparam(\latent_t)]$ and $\text{sign}[\valfuncLatent(\latent_t)], \forall t \in \{0,\hdots,T\}$ to check which latent states are in the failure set and unsafe set respectively. The two rows in Figure \ref{fig:sim-manip} correspond to each model's classification. We see that $\failureLatent$ correctly identifies that only the final observation at $t=28$ is in failure, since this is the only observation where the red blocks have fully fallen down. 
However, $\unsafeSetLatent$ detects that at timestep $t=14$, the robot has perturbed the blocks in such a way, that they are doomed to fail.

\para{Results: Quantitative} We rollout the un-shielded base $\dreamer$ policy and the policy shielded via $\ours$ and $\cmdp$ for 50 initial conditions of the blocks randomly initialized in front of the robot within $\pm \SI{0.05}{\meter}$ in the x and y directions.
We report the success, constraint violation, and incompletion rates in Table \ref{tab:sim_comparison}. 
We define a constraint violation as any rollout where at least one of the red blocks fall, success as any rollout where the robot successfully picked up the green block without toppling a red block, and incompletion as any rollout that does not violate constraints but failing to picking up the green block.
\begin{table}[h!]
\centering
\setlength{\tabcolsep}{5pt} 
\begin{tabular}{lccc}
\toprule
\textbf{Method} & \multirow{2}{*}{\textbf{\shortstack{Safe Success  \\  $\%$ ($\uparrow$)}}}& \multirow{2}{*}{\textbf{\shortstack{Constraint\\ Violation
$\%$ ($\downarrow$)}}}  & \multirow{2}{*}{\textbf{\shortstack{Incompletion  \\    $\%$ ($\downarrow$)}}} \\
& & \\ 
\midrule
\dreamer 
&   64    &  36 & \textbf{0}   \\
\cmdp (
$\epsilon_{\textrm{risk}} = 0.1$) &  68 &   28   & 4 \\
\cmdp (
$\epsilon_{\textrm{risk}} = 0.05$) &   8   &  22   & 70\\
\ours 
& \textbf{80}  &   \textbf{20} & \textbf{0}    \\
\bottomrule
\end{tabular}
\caption{\textbf{Visual Manipulation: Simulation.} Success at the task without any safety violations, constraint violations, and incompletion rates across 50 rollouts corresponding to 50 random initial conditions of the blocks. Task success is picking up the green block; constraint violation is where either of the red blocks fall down on the table.}
\label{tab:sim_comparison}
\end{table}

Despite the penalty for knocking over obstacles, we found that \dreamer learned to lift the block even when doing so would incur a safety violation. Although further tuning the reward function could potentially improve the behavior of the robot, reward engineering is notoriously tedious for engineers. This motivates using a safety filter to improve the safety of an unsafe task policy. We report the performance of \cmdp for two different values of $\epsilon^\textrm{risk}$ and found that \cmdp is extremely sensitive to choice of while $\epsilon^\textrm{risk}$, growing the task incompletion rate from $4\%$ when $\epsilon^\textrm{risk} = 0.1$ to $70\%$ when $\epsilon^\textrm{risk} = 0.05$ while only marginally improving safety. In contrast, \ours overrides our nominal task policy only when needed, significantly reducing the number of constraint violations while still succeeding at the task.

\section{Hardware Results: \\ Preventing Hard-to-Model Robot Failures}
\label{sec:hardware}
Finally, we design a set of experiments in hardware to see if our Latent Safety Filter can be applied in the real world (shown in Figure~\ref{fig:front-fig}). 
We use a fixed-base Franka Research 3 manipulator equipped with a 3D printed gripper from~\citep{chi2024universal}. 
The robot is tasked with interacting with an opened bag of Skittles on the table. 
The safety constraint is not to spill any Skittles. 
We test the efficacy of our \new{approach by deploying the \textit{same} Latent Safety Filter to safeguard a human teleoperator (Section~\ref{sec:human}) and a strong and weak Diffusion Policy \cite{chi2024diffusionpolicy} from spilling (Section~\ref{subsec:diffusion}), as well as stress-testing our safety filter to out-of-distribution candy bags and environment backgrounds (Section~\ref{subsec:ood}).}

\begin{figure*}[t!]
    \centering
    \includegraphics[width=\textwidth]{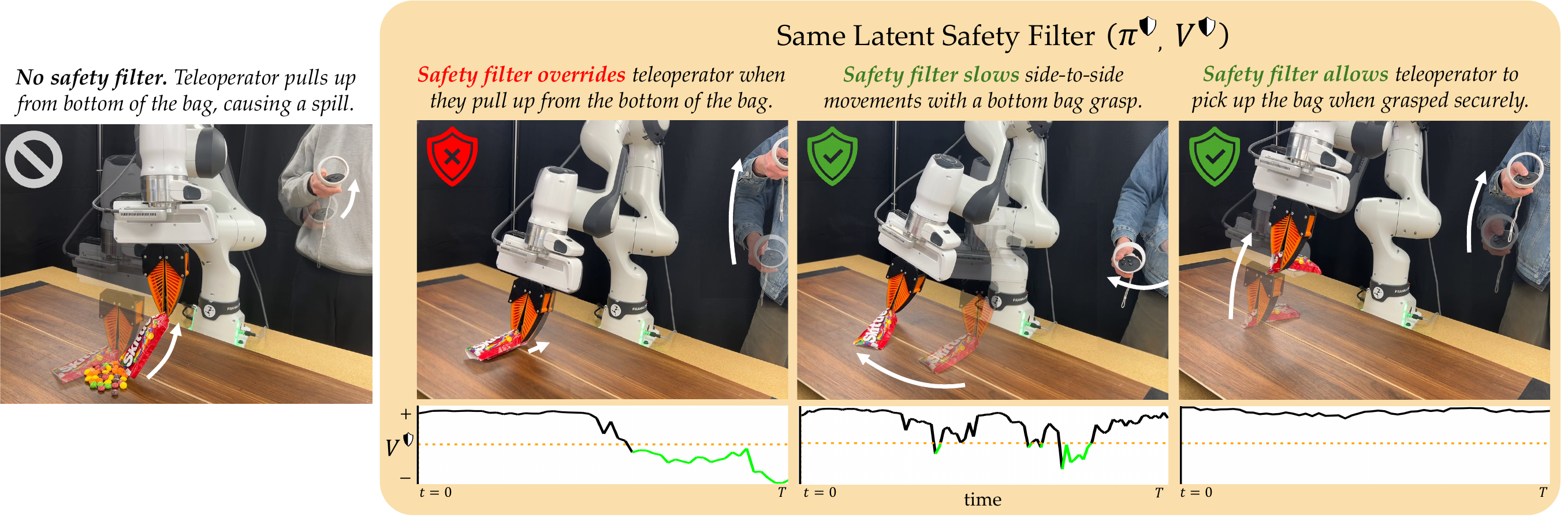}
    \caption{
    \textit{Far Left}: Without a safety filter, a teleoperator lifts the closed-end of the bag too quickly and spills the Skittles. \textit{Middle Left}: By using \ours, the same action of lifting the closed-end leads to the value function $\valfuncLatent^\shield$ dipping below the safe threshold (orange) and prompting the safety policy to override the teleoperator (green); the robot does not allow the human pull the bag up sharply. 
    \textit{Middle Right}: At the same time, \ours slows down the human's attempt to move the bag side-to-side while grasping the closed end, indicating that the safety filter has a nuanced understanding of which actions will and won't violate safety.
    \textit{Right}: Grasping the bag from the open end and lifting is deemed safe and is allowed by \ours.}
    \label{fig:teleop-results}
\end{figure*}

\para{Safety Specification} Our safety specification is to prevent the contents of the Skittles bag from falling out of the bag. 
Given only image observations and proprioception, this problem is clearly partially observed since the robot cannot directly recover the position of the Skittles in the bag.
Even if privileged state information were available, designing a function to characterize the set of failure states and a dynamics model for interactions between all relevant objects (e.g., the manipulator, the soft bag, the Skittles, and the table) would be extremely difficult. 

\para{Latent Safety Filter Setup} We use DINO-WM \cite{zhou2024dinowm}, a Vision Transformer-based world model that uses DinoV2 as an encoder \cite{oquab2023dinov2}. 
The manipulator uses a 3rd person camera and a wrist-mounted camera and records $3 \times 256 \times 256$ RGB images at 15 Hz.
For world model training, we collected a dataset  $\mathcal{D}_\text{WM}$ of 1,300 offline trajectories: 1,000 of the trajectories are generated sampling random actions drawn from a Gaussian distribution at each time step, 150 trajectories are demonstrations where the bag is grasped without spilling any Skittles, and 150 demonstrations pick up the bag while spilling candy on the table. 
We manually labeled the observations in the trajectory dataset for apparent failures. 
However, in principle, we believe that this data annotation process could also be automated using alternative methods, like state-of-the-art foundation models (e.g., vision-language models). 

Our world model is trained by first preprocessing and encoding the two camera view using DINOv2 to obtain a set of dense patch tokens for each image. 
We use the DINOv2 ViT-S, the smallest DINOv2 model with 14M parameters, resulting in latent states $\latent$ of size $256 \times 384$ corresponding to 256 image patches each with embedding dimension 384. 
The transition function is implemented as a vision transformer, which takes as input the past $H=3$ patch tokens, proprioception, and actions to predict the latent. The transformer employs frame-level causal attention to ensure that predictions can only depend on previous observations. 
The model is trained via teacher-forcing minimizing mean-squared error between the ground-truth DINO embeddings of observations and proprioception information from $\mathcal{D}_\text{WM}$ and the embeddings and proprioception predicted by the model. 
Additional details on the model and hyperparameters are included in the appendix and in \cite{zhou2024dinowm}. 
After world model training, we separately train the failure classifier (implemented as a 2-layer MLP with hidden dimension 788 and ReLU activations) on the DINO patch tokens corresponding to the manually labeled constraint-violating observations. 
For approximating the HJ value function, we use DDPG \cite{lillicrap2019continuouscontroldeepreinforcement, li2025certifiabledeeplearningreachability}.


\subsection{Shielding Human Teleoperators}
\label{sec:human}

To emphasize the policy-agnostic nature of our latent safety filters, we demonstrate filtering the actions of a teleoperator.

\para{Setup} The teleoperator controls the end-effector position and gripper state via a Meta Quest pro similar to \cite{khazatsky2024droid}, and can freely move the robot around. They are tasked with interacting with the Skittles bag however they like. The safety filter operates according to the following control law:
\begin{align}
    \action^\text{exec}_{t} = 
    \begin{cases}
    \policyTask(\latent_t) ,& \text{if } \valfunc(\hat{\latent}_{t+1}) > \epsilon\\
    \fallbackLatent(\latent_t),              & \textrm{otherwise}  \\ 
    \end{cases}
\end{align}
where $\hat{\latent}_{t+1} \sim p_\dynparam(\hat{\latent}_{t+1} \mid \latent_t,  \policyTask(\latent_t))$ is a one-step rollout of the world model using the action proposed by the unshielded \new{teleoperator}.  \new{In hardware experiments, we set $\epsilon=0.3$.} Both the teleoperation and safety filtering were executed at $15$ Hz.

\para{Results: Shielding Unsafe Grasps and Dynamic Motions} 
We visualize our qualitative results in Figure \ref{fig:teleop-results}. 
Un-shielded by our safety filter, the teleoperator can grab the opened bag of Skittles by the base and pull up sharply, spilling its contents on the table (left, Figure~\ref{fig:teleop-results}). 
By using \ours, the same behavior gets automatically overridden by the safety filter, preventing the teleoperators ``pull up'' motion from being executed and keeping the Skittles inside (center, Figure~\ref{fig:teleop-results}). 
At the same time, the latent safety filter is not overly pessimistic (right-most images in Figure~\ref{fig:teleop-results}). 
When the teleoperator moves the Skittles bag side-to-side while grasping the \textit{bottom} of the opened bag, the safety filter accurately accounts for these dynamics and minimally modifies the teleoperator to slow them down, preventing any Skittles from falling out while still allowing the general motion to be executed. 
When the teleoperator chooses a safe grasp---grabbing the bag by the top, open side---the safety filter does not activate and allows the person to complete the task safely and autonomously. 

\subsection{Shielding \new{Autonomous} Imitation-Learned Policies}
\label{subsec:diffusion}

\new{Next we study how well the same Latent Safety Filter from Section~\ref{sec:human} can shield autonomous imitation-learned (IL) policies. 
Specifically, we test whether the latent safety filter does \textit{not} impede a strong IL policy (i.e., our filter is not overly conservative) and \textit{ improves} the safety of a suboptimal IL policy (i.e., our filter shields effectively), while removing teleoperator bias that may be present in our prior experiments.}

\para{Methods} \new{For our base task policy, $\policyTask(\obs)$, we use a generative imitation-learned (IL) policy trained with a diffusion objective \cite{chi2024diffusionpolicy} and which takes as input RGB images and end effector pose as observations $\obs \in \obsSpace$ (implementation details can be found in the Appendix). 
We train two diffusion policies---\diffpolicyadv and \diffpolicyopt---which represent relevant extremes of a base policy's capabilities. 
\diffpolicyopt represents the ``upper bound'' of a strong base policy that uses carefully curated demos of the task. 
We use this baseline to study whether our safety filter is not overly conservative when shielding a strong base policy.  
We train it with 100 teleoperated demonstrations 
wherein the expert grasps the Skittles bag from the middle and lifts it off the table without spilling. 
We also train \diffpolicyadv, which represents a ``lower bound'' of a base policy trained with demonstrations that could lead to unsafe outcomes. 
This policy is trained with 100 potentially unsafe teleoperated 
demonstrations: the expert grasps and lifts the Skittles bag from its closed end, but the opening is internally sealed during data collection time so that no Skittles could fall out. 
This results in a base policy that has an incomplete understanding of how to interact safely with the Skittles bag, allowing us to test our safety filter's ability to prevent failures in a controlled and repeatable manner. 
}

\begin{figure}[t!]
    \centering
    \includegraphics[width=\linewidth]{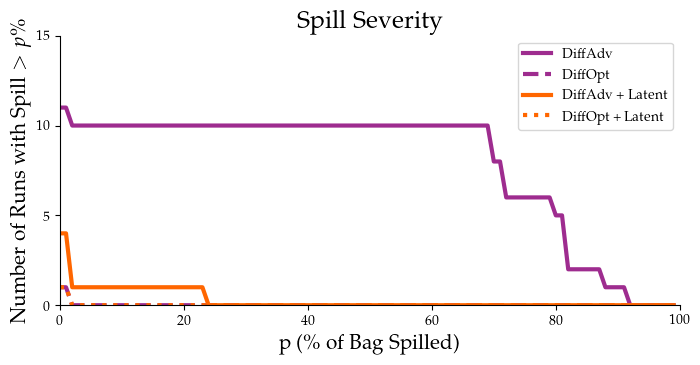}
    \caption{\textbf{Shielding IL Policies: Hardware Results.} 
    Percent of bag spilled ($p$) vs number of runs that spilled at least $p\%$ of the bag.
    While \diffpolicyadv frequently spills a large percentage of the bag ($\sim 85\%$), \diffpolicyadv + \ours spills less than $5\%$ of the bag in all but one of the constraint-violating rollouts. \new{\diffpolicyopt with and without \ours spills only 1 skittle in across all 15 rollouts, showing that latent safety is not overly conservative.}}
    \label{fig:hardware-hist}
    \vspace{-0.5cm}
\end{figure}

\new{\para{Metrics}
We compare the performance of the \new{$\policyTask \in \{\diffpolicyadv, \diffpolicyopt\}$ with and without using \ours (yielding four methods in total). 
We use exactly the same \textit{Latent Safety Filter} as we used to shield the human teleoperator in Section \ref{sec:human}.}
For each method, we record 15 rollouts where the policy successfully grasped the bag (ignoring missed grasps) 15 times in hardware. We measure the frequency of constraint violations (if even one Skittle falls out during an episode) and spill severity (percentage of the Skittles spilled) \new{in each of these trajectories.}}

\para{Results: Shielding a Weak IL Policy} 
\new{When \ours shields the weak \diffpolicyadv, we see a 63.6\% decrease in constraint violations compared to \diffpolicyadv acting alone.} 
While \ours does still fail $26.4\%$ of the time, we note that 3 out of the 4 failures only spilled a single skittle. 
In contrast, \diffpolicyadv's autonomous spill rate was $73.4\%$ with many instances where a large percentage of the bag was spilled.
To better understand the \textit{severity} of the constraint violations, we report in Figure~\ref{fig:hardware-hist} how often each method spilled more than $p\%$ of the bag.
While \diffpolicyadv frequently spills a large percentage of the bag ($\sim 85\%$), the safety filter spills less than $5\%$ of the bag in all but one of the constraint-violating rollouts, where it spilled only \new{$24\%$}. \new{Overall, \ours minimizes both the failure rate and severity when the base IL policy is erroneous and can cause difficult-to-model failures.}

\new{\para{Results: Shielding a Strong IL Policy} 
We report the severity of constraint violations in Figure~\ref{fig:hardware-hist}.
We note that both when \diffpolicyopt acts alone and when it is shielded by \ours, the policies safely grasp and lift the bag in the first 15 trials. Both methods exhibited only one safety violation, where a single Skittle was spilled. 
Overall, we see that our Latent Safety Filter is not overly-conservative, allowing a strong base policy to operate without unnecessary overrides. 
We also note that practically, since the same Latent Safety Filter was used for both the weak and the strong base IL policy, this provides a promising avenue for safely improving a base task-driven policy without the need to also change the safety representation and fallback controller.
}



\new{\subsection{Testing Out-of-Distribution Generalization of Latent Safety}
\label{subsec:ood}

Finally, we stress-test the performance of the same Latent Safety Filter on out-of-distribution (OOD) bag colors, candy dynamics, and background changes. 

\para{Setup}  Recall that our Latent Safety Filter was trained only on classic red Skittles bag and with a wooden table-top background. In these tests, we first vary the color of the Skittles bag, where OOD Skittles are green (Sour) and purple (Wild Berry). We also test OOD candy bags: M\&Ms Classic, M\&Ms Peanut Butter, M\&Ms Peanut. Note that these are both visually and dynamically OOD: we qualitatively observe that M\&Ms bags are stiffer and have a papery texture compared to the Skittles bags. Furthermore, the M\&Ms are have differing weights (e.g., with peanuts). Finally, we also deploy our method to interact with a red Skittles bag, but cover the table with an OOD black cloth.

\para{Methods} To isolate the influence of OOD conditions on our Latent Safety Filter, we replay a known unsafe demonstration from our dataset in open-loop as our task ``policy''. We reset the bag to the same initial condition and shield this replayed demonstration with \ours filter for all OOD conditions.

\para{Results} 
We plot the safety value function and the final observation of the system for the OOD Skittles and background in Figure \ref{fig:skittle-ood} and OOD M\&Ms in Figure \ref{fig:mm-ood}.  
When shielding OOD Skittles bags and the novel background, the filtering override profile is similar for all of the bag colors and results in the same final observation where none of the candy is spilled. 
In contrast, we observe a noticeably different safety value profile for the OOD M$\&$Ms in Figure \ref{fig:mm-ood}. For both the Classic and Peanut M$\&$Ms, our method fails to prevent spills. 
Note that for the Classic  M$\&$Ms (brown), the safety filter started to activate prior to the candy spilling (grey line) but was unable to prevent a spill. We hypothesize that the powerful, pre-trained DINOv2 embeddings are able to identify semantic equivalences between the differing bags\footnote{\new{In the Appendix, we provide qualitative evidence by visualizing the top three PCA components of the DINOv2 embeddings of all OOD scenarios.}}, which would be sufficient for safe control across dynamically equivalent bags. However, despite the encoder's ability to map visually similar observations to similar embeddings, there is no reason to suggest that the transition model can generalize across different bag dynamics in our extremely low data regime, which would explain the gap in closed-loop performance between M\&Ms and Skittles.  
\begin{figure}[h!]
    \centering
    \hspace{-0.75cm}
    \includegraphics[width=1.05\linewidth]{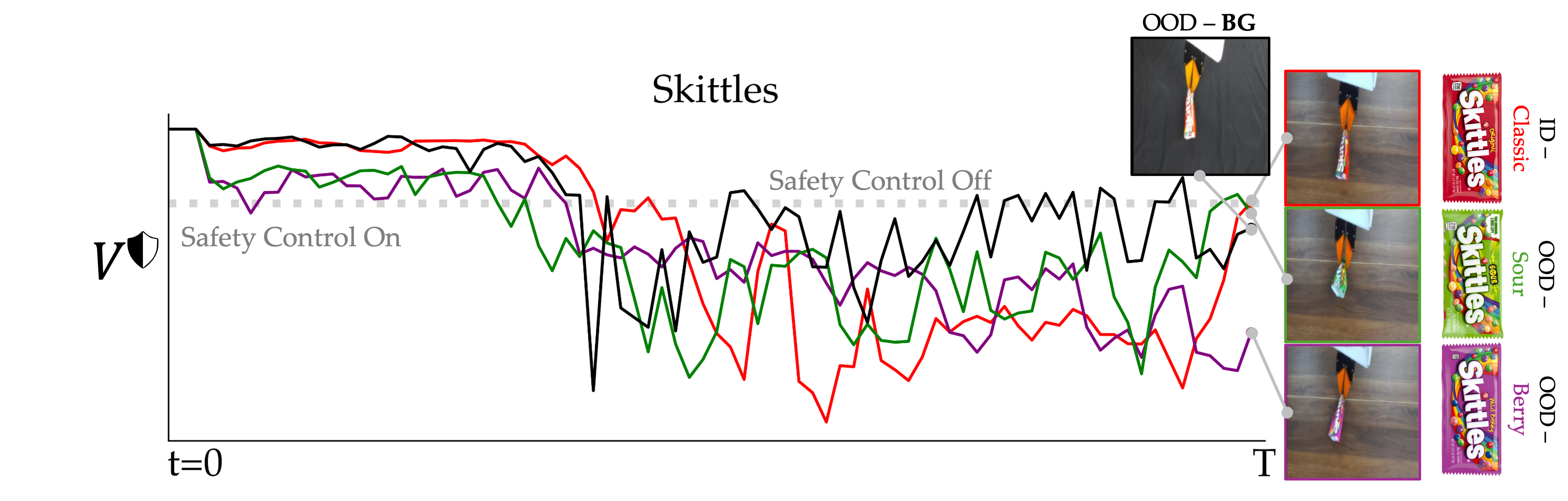}
    \caption{\new{\textbf{OOD Generalization: Skittles.} 
    Our latent safety filter is trained only on a red Skittles bag. It is deployed to shield an open-loop known unsafe trajectory for two OOD skittles bag colors and an OOD background. \ours generalizes---maintaining the same performance of preventing spills---to OOD Skittles bag colors and OOD background change.}}
    \label{fig:skittle-ood}
\end{figure}
\begin{figure}[h!]
    \centering
    \hspace{-0.75cm}
    \includegraphics[width=1.05\linewidth]{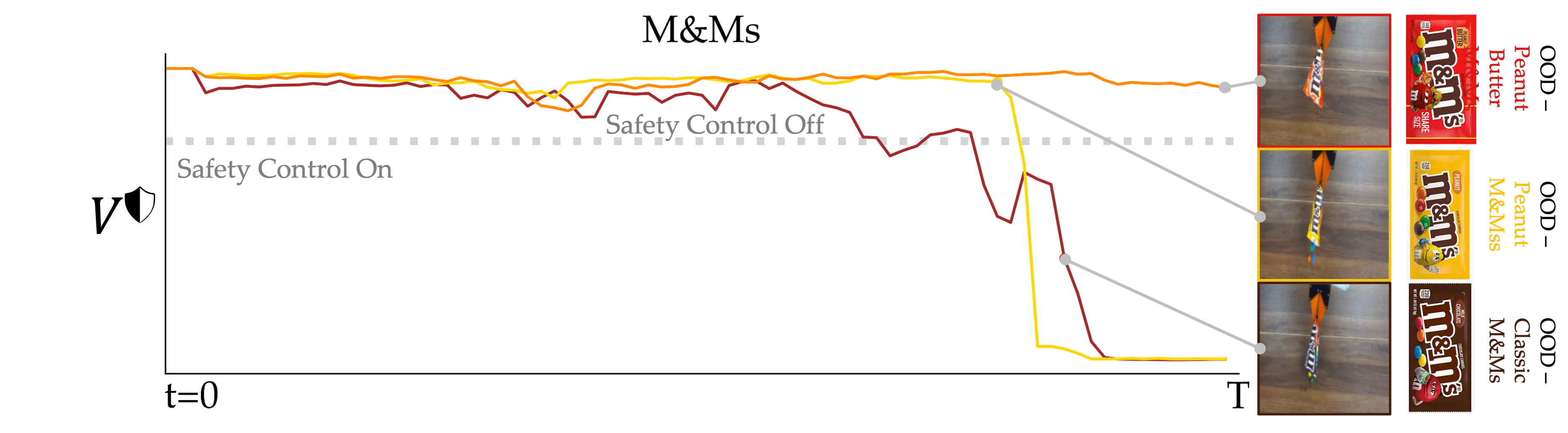}
    \caption{\new{\textbf{OOD Generalization: M\&Ms.} Our latent safety filter is trained only on a red Skittles bag. It is deployed to shield an open-loop known unsafe trajectory. \ours is deployed with 3 M\&M bags with different colors and dynamics. Our filter does \textit{not} prevent the manipulator from lifting these OOD bags. For the brown bag, even though the filter begins to override the recorded trajectory, it does not manage to prevent the spill, potentially due to differing dynamics.}}
    \label{fig:mm-ood}
\end{figure}

}
\section{Related Work}


\para{Safety Filtering for Robotics} 
Safety filtering is a control-theoretic approach for ensuring the safety of a robotic system in a way that is agnostic to a task-driven base policy \cite{hsu2023safety}. 
A safety filter monitors a base task policy and overrides it with a safe control action if the system is on the verge of becoming unsafe. Control barrier functions \cite{ames2019control}, HJ reachability \cite{margellos2011hamilton, mitchell2005time, Fisac15}, and model-predictive shielding \cite{wabersich2022predictive} are all common ways of instantiating safety filters (see \cite{wabersich2023data} for a survey). 
The most relevant recent developments include computing safety filters from simulated rollouts of first-principles dynamics models or high-fidelity simulators via reinforcement learning or self-supervised learning \cite{hsu2023isaacs, hsu2021safety, fisac2019bridging, bansal2021deepreach, robey2020learning}, safety filters that operate on a belief-state instead of a perfect state \cite{ahmadi2019safe, bajcsy2021analyzing, hu2023deception, vahs2023belief}, and safety filters that keep a system in-distribution of a learned embedding space \cite{castaneda2023distribution}. 
Our work contributes to the relatively small body of work that attempts to bridge the gap between high-dimensional observations, such as LiDAR \cite{lin2024filterdeployallrobust, AgileButSafe} or RGB images \cite{tong2023enforcing, hsuren2022slr}, and safety filters. 
However, unlike previous work, our method does not infer hand-crafted intermediary representations of state from the high-dimensional observations nor does it represent safety as collision-avoidance. 
Instead, our safety filter operates in the latent space of a world model, reasoning directly about the embedded RGB observations and shielding against hard-to-model failure specifications.


\para{Latent Space Control} 
The \new{control and} model-based reinforcement learning community has recently demonstrated the potential for using generative world models for real-world robot control \cite{mendonca2023structured, wu2022daydreamer}. 
One of the advantages of world models is that they transform a control problem with partial observability into a Markov decision process in the learned latent state space. 
Many methods for shaping this latent state representation exist, from observation reconstruction \cite{hafner2019learning, hafner2020dreamerv2, hafner2024masteringdiversedomainsworld}, teacher-forcing \cite{zhou2024dinowm}, reward predictions \cite{Hansen2022tdmpc, hansen2024tdmpc2}\new{, or explicit metric learning in the latent space \cite{hong2022dynamics}}. 
While prior works traditionally use the world model to learn a policy for a specific task, we use the world model to learn a policy-agnostic safety filter that reasons about unsafe consequences it can ``imagine'' (but are hard to model) in the latent space \new{ via reachability analysis. 
We note that our paper is not the first to apply control-theoretic principles in the latent space of a learned world model. 
Latent models have been used to estimate the regions of attraction for data-driven policies \cite{vieira2024morals}, collision-free motion planning \cite{ichter2019robot}, and forward reachability methods have been used for structured exploration toward states that are far from a system's initial state \cite{bharadhwaj2021leaf}. 
Instead, we take a \textit{backward} reachability approach \cite{bansal2017hamilton} that determines the set of states that will \textit{inevitably} lead to failure despite the robot's best effort to prevent failure. 
This allows us to simultaneously compute a runtime monitor as well as a safety recovery policy that can preemptively prevent failures.}

\new{\para{Learning about Safety from Expert Demonstrations}} 
\new{Given an offline dataset of expert data and a privileged state space, there are two predominant methods for learning about safety. 
The first uses expert demonstrations to learn a safe policy directly. 
For example, control barrier functions can be learned from expert demonstrations if the ground-truth dynamics and their Lipschitz constants are known \cite{lindemann2024learning} and \cite{robey2020learning}. 
Inverse constraint learning methods
~\cite{scobee2019maximum, chou2020learning, kim2024learning} infer constraints given expert trajectories that are optimal with respect to a known reward and unknown constraint; high-reward trajectories that the expert demonstrator did not take must have a safety violation. 
The assumption of a known reward has been relaxed in \cite{chou2020learninguncertainty, lindner2024learning} but still requires trajectories that are (locally) optimal with respect to some task. 
Our work does not assume expert demonstration trajectories; it only assumes observations labeled with a binary indication of failure (which we use to learn our failure state classifier). 
We also do not assume a known state or dynamical system model, can use diverse robot deployment data regardless of its optimality (for world model building and failure classifier training), and we compute the safety policy as an optimization rather than requiring demonstrations of safety-preserving behavior. 
}

\new{\para{Computing Safety Behaviors from Offline Data} Our method aligns with the other set of approaches focusing} on programmatically computing the safety filter via synthesis techniques (e.g., optimal control) once the \textit{safety specification} (i.e., the failure set) is encoded or learned.
\new{Constraints more nuanced than collision avoidance can be specified with signal and linear temporal logic using demonstrations \cite{bartocci2022survey} or language \cite{finucane2010ltlmop}, but have still been restricted to constraints that can be expressed using hand-designed predicates of privileged state variables.}
\new{Prior works have learned about failure states by using intervention data in a learned latent space \cite{liu2023modelbasedruntimemonitoringinteractive, liumulti} or binary indicators of constraint violation (provided by an oracle) \cite{srinivasan2020learning, thananjeyan2021recovery, wilcox2022ls3}. Our method leverages labels on robot observations (provided by an annotator) to learn a margin function on the embeddings of high-dimensional observations, which represents hard-to-model constraints via its zero-sublevel set.}
We then utilize reinforcement-learning-based HJ reachability to synthesize safe behavior automatically, rather than relying on a demonstrator to provide recovery behavior as is done in \cite{liu2023modelbasedruntimemonitoringinteractive,lindemann2024learning, robey2020learning}.
\new{
While similar to our work in its use of a latent space, \cite{wilcox2022ls3} solves for a policy which co-optimizes safety and task performance. This renders the resulting safe policy \textit{task-specific}, preventing its use as a general-purpose, policy-agnostic safety filter like we formulate. Furthermore, while prior work uses additional interactions with the ground truth environment to update its safety policy and world model \cite{liu2023modelbasedruntimemonitoringinteractive, wilcox2022ls3, thananjeyan2021recovery}, our work learns a safety filter entirely within a frozen world model, minimizing the need for any additional (potentially unsafe) environment interactions.


}

\section{Limitations}
Our latent HJ reachability formulation generalizes the space of failures robots can safeguard against.
However, it is not without its limitations. 
One limitation is that the safety filter can only protect against outcomes that the world model can predict.
This means that the world model needs to be trained on some amount of unsafe data in order to effectively predict these unsafe outcomes and compute control policies that steer clear of them. 
Additionally, the least-restrictive safety filter mechanism we implemented returns a single control action that attempts to steer the robot away from danger maximally and at the last moment. 
More sophisticated approaches to search through the set of safe actions and align them with the base policy's task performance should be explored  \cite{hsu2023safety, wabersich2023data}. \new{We also inherit some limitations from value-based safe control techniques, such as the need to recompute a safety value function for any new additional constraints. 
However, we view this as an exciting opportunity for future work where the safety value function is parameterized by the failure specification \cite{borquez_parameter-conditioned_2023} for additional flexibility without the need for recomputation.} 
Finally, since we are concerned with robot safety, it is important to acknowledge what types of assurances we can expect from this framework. 
Although our method is grounded in rigorous theory, our practical implementation currently lacks formal assurances due to the combination of possible errors when learning the world model, failure classifier, and resulting HJ value function. Characterizing how the errors in one learned component propagate to the downstream safety assurances is exciting future work.

\section{Conclusion}
In this work, our goal was to generalize robot safety beyond collision-avoidance, accounting for hard-to-model failures like spills, items breaking, or items toppling. 
We introduce \textit{Latent Safety Filters}, a generalization of the safety filtering paradigm that operates in the learned representation of a generative world model. We instantiated our method on a suite of simulation and hardware experiments, demonstrating that our latent reachability formulation is comparable to privileged state formulations, outperforms other safe control paradigms while being less sensitive to hyperparameter selection, and protects against extremely hard-to-specify failures like spills in the real world for both generative IL policies and human teleoperation. 
Future work should thoroughly investigate the uncertainties within each component of our latent space safety generalization and investigate theoretical or statistical assurances on this new safety paradigm. 

\section{Acknowledgements} 
\noindent The authors would like to thank Gaoyue Zhou for the guidance on DINO-WM and Chenfeng Xu for assistance with ViTs. We also thank Gokul Swamy, Yilin Wu, and Junwon Seo for helpful discussions. This material is supported in part by the National Science Foundation Graduate
Research Fellowship Program under Grant 
 NoDGE2140739 and NSF CAREER Award No. 2441014. Any opinions,
findings, and conclusions or recommendations expressed in this material are those of the author(s) and do not necessarily reflect the views of the National Science Foundation.


\bibliographystyle{plainnat}
\bibliography{references.bib}

\newpage 
\clearpage
\appendix

\subsection{Hyperparameters}

In Section \ref{sec:sim}, we showcase results for a Recurrent State Space Model~(RSSM) \cite{hafner2019learning} that uses both a deterministic recurrent state $h_t$ and stochastic component $x_t$.
\begin{equation}
\begin{aligned}
    &h_{t+1} = f_{\dynparam}(h_{t}, x_{t}, a_{t}) \;\;\;\;\;\;\;
    \hat{x}_t = p_\dynparam(\hat{x}_t \mid h_t) \\
        &x_{t} = \enc_\dynparam(x_{t} \mid h_t, o_t) \;\;\;\;\;\;\;\;\;\;\;\; \hat{o}_t = \text{dec}_\dynparam(\hat{o}_t \mid h_t, x_t)
\end{aligned}
 \end{equation}
 For the RSSM we define $\latent := \{h_t, x_t \}$.
 The RSSM shares weights between each module and is trained to minimize the KL divergence between the encoding of the current latent $x$ and the latent $\hat{x}_t$ predicted by the transition dynamics. The RSSM additionally utilizes an observation reconstruction objective to maintain the informativeness of its latent state. We refer the readers to \cite{hafner2019learning, hafner2020dreamerv2, hafner2024masteringdiversedomainsworld} for a more comprehensive overview of RSSMs. We use the implementation from \cite{dreamerv3-torch} and list relevant hyperparameters in Table \ref{tab:dreamer}. For Dubin's car experiments, we use a continuous stochastic latent parameterized as a 32-dimensional Gaussian. We use 32 categorical random variables with 32 classes for the high-dimensional simulation experiments.
\begin{table}[h]
    \centering
    \begin{tabular}{l l}
        \toprule
        \textbf{Hyperparameter} & \textbf{Values}  \\
        \midrule
        Image size & 128 \\
        Optimizer & Adam \\
        Learning rate (lr) & $1e-4$ \\
        Hidden dim & 512 \\ 
        Dyn deterministic  & 512 \\
        Activation fn & SiLU \\
        CNN depth & 32 \\
        Batch size & 16 \\
        Batch Length & 64 \\
        Recon loss scale & 1 \\
        Dyn loss scale & 0.5 \\
        Representation loss scale & 0.1 \\ 
    \bottomrule
    \end{tabular}
    \caption{Hyperparameters for Dreamer}
    \label{tab:dreamer}
\end{table}

In Section \ref{sec:hardware} we train DINO-WM~\cite{zhou2024dinowm}, a transformer-based world model that uses the patch-tokens of DINOv2~\cite{oquab2023dinov2} as a representation for the latent state. Here, the DINOv2 encoder is kept frozen, and we train only the parameters of a vision transformer, which is used to predict future patch tokens $\hat{\latent}_{t+1}$ conditioned on a sequence of actions $a_{t-H:t}$ and tokens $\latent_{t-H:t}$ from the previous $H$ timesteps.

\begin{equation}
\begin{aligned}
    &z_{t} = \enc_\encparam(o_t) \;\;\;\;\;\; \hat{\latent}_{t+1} = p_{\dynparam}(\latent_{t-H:t}, \action_{t-H:t})
\end{aligned}
\end{equation}
This model is trained via teacher-forcing with the following consistency loss
$\mathcal{L}(\dynparam) = || \enc_\encparam(o_{t+1}) - p_{\dynparam}(z_{t-H:t}, a_{t-H:t})||$. We tokenize both the wrist and front camera views, leading to two sets of image patches with embedding dimension 384. We additionally encode the action and proprioception into a 10-dimensional latent vector. We concatenate the image patches for both cameras, action embedding, and proprioception embedding for $H=3$ frames. We use learnable positional and temporal embeddings. We pass the output from the transformer into a wrist camera, front camera, and proprioception head, which predicts the corresponding input for the next time step. These are implemented as MLP heads with three layers, a hidden dimension of $788$, and a learning rate of $5e-5$. The remaining hyperparameters are the same as \cite{zhou2024dinowm} and are showing in Table \ref{tab:dinowm}
\begin{table}[h]
    \centering
    \begin{tabular}{l l}
        \toprule
        \textbf{Hyperparameter} & \textbf{Values}  \\
        \midrule
        Image size & 224 \\
        DINOv2 patch size & $(14 \times 14, 384)$ \\ 
        Optimizer & AdamW \\
        Predictor lr & 5e-5  \\
        Decoder lr & 3e-4 \\
        Action Encoder lr & 5e-4 \\
        Action emb dim & 10 \\
        Proprioception emb dim & 10 \\
        Batch size & 16 \\
        Training iterations & 100000 \\
        ViT depth & 6 \\
        ViT attention heads & 16 \\
        ViT MLP dim & 2048 \\
    \bottomrule
    \end{tabular}
    \caption{Hyperparameters for DINO-WM}
    \label{tab:dinowm}
\end{table}

The Dubin's car experiments use a discrete action space, so we train the value function using DDQN \cite{van2016deep} using the toolbox from \cite{hsu2021safety}. The Q-function is implemented as a 3-layer MLP with 100-d hidden dimension. The remaining hyperparameters are shown in Table \ref{tab:rarl}.
\begin{table}[h]
    \centering
    \begin{tabular}{l l}
        \toprule
        \textbf{Hyperparameter} & \textbf{Values}  \\
        \midrule
        Optimizer & AdamW \\
        Learning rate & 1e-3  \\
        Learning rate decay & 0.8 \\
        Hidden dims & $[100, 100]$ \\
        Time discount $\gamma $ & 0.9999 \\
        Activations & Tanh \\
        Batch size & 64 \\
        Training iterations & 400000 \\ 
    \bottomrule
    \end{tabular}
    \caption{Hyperparameters for DDQN HJ Reachability}
    \label{tab:rarl}
\end{table}

For both the simulation and hardware manipulation experiments, we use DDPG \cite{lillicrap2019continuouscontroldeepreinforcement} using the implementation from \cite{li2025certifiabledeeplearningreachability} using the standard discounted Bellman equation from \cite{fisac2019bridging}.

\begin{table}[h]
    \centering
    \begin{tabular}{l l}
        \toprule
        \textbf{Hyperparameter} & \textbf{Values}  \\
        \midrule
        Optimizer & AdamW \\
        Actor lr & 1e-4  \\
        Critic lr & 1e-3 \\
        Actor + Critic hidden dims & $[512, 512, 512, 512]$ \\
        Time discount $\gamma $ & 0.9999 \\
        Activations & ReLU \\
        Batch size & 512 \\
        Epochs & 50 \\ 
    \bottomrule
    \end{tabular}
    \caption{Hyperparameters for DDPG HJ Reachability}
    \label{tab:lcrl}
\end{table}
\begin{figure*}[ht!]
    \centering
    \includegraphics[width=.95\linewidth]{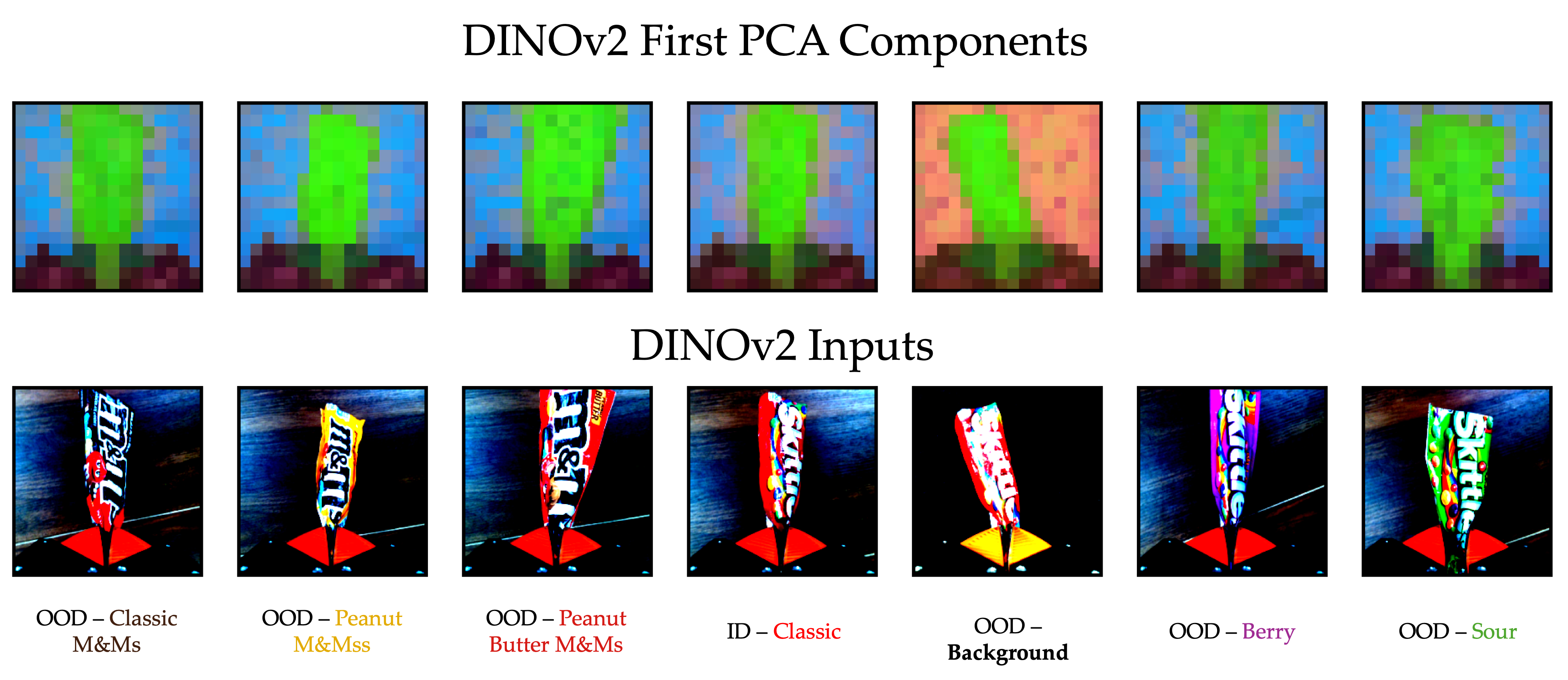}
    \caption{\new{Bottom: Input to DINOv2 after applying image transformations. Top: Top 3 PCA components visualized in the RGB channels. Despite the difference in bag or background color, the PCA visualization shows that all bags are similarly represented in the DINOv2 embedding space.}}
    \label{fig:DINO}
\end{figure*}
To smooth the loss landscape for reachability learning, we apply $\tanh$ to the output of $\marginfunc_\ellparam$. This is because the loss function for the failure classifier may output very different values to states that are similarly constraint violating / far from constraint violation, so long as it achieves low loss by \eqref{eq:failure-classifier-loss}. This thresholding ensures that sufficiently safe or unsafe states are evaluated similarly by $\tanh(\marginfunc_\ellparam(\latent))$.

\new{
For the Diffusion Policy experiments, we match the CNN-based implementation from \cite{chi2024diffusionpolicy} using the hyperparameters shown in Table \ref{tab:dp}. While we train for 500000 iterations, we in practice choose the best-performing checkpoint for each model.
\begin{table}[h]
    \centering
    \begin{tabular}{l l}
        \toprule
        \textbf{Hyperparameter} & \textbf{Values}  \\
        \midrule
        State Normalization & Yes \\
        Action Normalization & Yes  \\
        Action Space & End Effector Delta Position \\
        Rotation Representation & Axis Angle \\
        Action Chunk & 16 \\
        Image Chunk & 2 \\
        Image Size & 256 \\
        Batch size & 100 \\
        Training Iterations & 500000 \\ 
        Learning Rate & 1e-4 \\
        Learning Rate Schedule & Cosine \\
        Optimizer & AdamW \\
    \bottomrule
    \end{tabular}
    \caption{\new{Hyperparameters for Diffusion Policy}}
    \label{tab:dp}
\end{table}
}
\subsection{Common Questions}

\ques{How do we get labels for training $\failureLatent$?}

When training in simulation, we used privileged data provided by the simulator. For the real-world hardware tasks, we labeled the observations from the world model training trajectories by hand. In hardware, for our 1,300 collected trajectories, this simply meant identifying a single frame in the trajectory where all subsequent timesteps were in violation of the safety constraint and all previous timesteps were safe. This took the lead author about 2.5 hours. While this process is manually intensive right now, it is (1) easier for non-experts to label observations that appear in failure instead of designing a functional representation of the failure and (2) in principle, foundation models (e.g., VLMs) could be used to annotate visually-apparent failures automatically. Furthermore, when training the failure classifier separately from the world model, one can select a smaller subset of data to label and train the failure classifier.

\new{\ques{How does the policy generalize?}

While an in-depth study on the generalization capabilities of world models and resulting policies is out of the scope of this work, we provide qualitative evidence to our claim in the main body that the DINOv2 \cite{oquab2023dinov2} encoder used in DINO-WM \cite{zhou2024dinowm} may help generalize across semantically equivalent instances (e.g., bags of different colors). In Figure \ref{fig:DINO} follow the visualization procedure from \cite{oquab2023dinov2} and compute a principal component analysis (PCA) between all of the image patches, and show their first three components as a channel in an RGB image. Despite the difference in bag and/or background color, the PCA visualization of the DINOv2 features shows that the bags of candy are consistently represented with the color green. Although this does not reveal information beyond the first three PCA components, it suggests that the DINOv2 encoder maps these images to similar embeddings.


}

\end{document}